\def\year{2019}\relax
\documentclass[letterpaper]{article} 
\usepackage{aaai19}  
\usepackage{times}  
\usepackage{helvet}  
\usepackage{courier}  
\usepackage{url}  
\usepackage{graphicx}  
\usepackage{amsmath}
\usepackage{amssymb}
\usepackage{pifont}
\newcommand{\cmark}{\ding{51}}%
\newcommand{\xmark}{\ding{55}}%
\usepackage{wasysym}
\usepackage{float}
\usepackage{dblfloatfix}
\usepackage{subfigure}
\usepackage[dvipsnames]{xcolor}
\usepackage{url}
\usepackage{graphicx}
\usepackage{amsfonts}
\usepackage{booktabs}
\usepackage{multirow}
\usepackage{multicol}
\usepackage{dashrule}
\usepackage{adjustbox}
\usepackage{tabularx} 
\usepackage{colortbl}

\usepackage{CJKutf8}

\usepackage{cleveref} 

\crefformat{section}{\S#2#1#3}
\crefformat{subsection}{\S#2#1#3}
\crefformat{subsubsection}{\S#2#1#3}
\crefrangeformat{section}{\S\S#3#1#4 to~#5#2#6}
\crefmultiformat{section}{\S\S#2#1#3}{ and~#2#1#3}{, #2#1#3}{ and~#2#1#3}

\begin{document}
%
\title{Meta-path Augmented Response Generation}
\author{Yanran Li \and Wenjie Li\\
Department of Computing, The Hong Kong Polytechnic University\\
\{csyli,cswjli\}@comp.polyu.edu.hk\\
}
\maketitle
\begin{abstract}
We propose a chatbot, namely \textsc{Mocha} to make good use of relevant entities when generating responses. Augmented with meta-path information, \textsc{Mocha} is able to mention proper entities following the conversation flow.
\end{abstract}

\section{Introduction}

Recent years have witnessed a rise in building automatic chit-chat conversational systems (a.k.a. chatbots). To generate informative responses, these chatbots need to be aware of conversation-related knowledge. For example, when talking about a film, it is natural to mention its director and actors.

We thus endow the chatbots with the entity reasoning mechanism, i.e., the ability to mention proper entities with reference to an associated knowledge base (KB) when generating responses. To ensure response quality, the generated entities should not only be relevant, but also coherent to the conversation flow. When previous conversation had focused on the actors of a film, it would be weird to suddenly mention the film's writer. Inspired by~\cite{dong2017metapath2vec}, we capture conversation flow using meta-paths over the mentioned entities. A meta-path is a sequence of object types modeling a particular semantic relationship. An example meta-path of entity mentions is \emph{actor}$\to$\emph{film}$\to$\emph{film}. Since meta-path information greatly reflects the conversation flow, it is better to mention entities following the meta-path when generating responses.

In this work, we develop a \textbf{M}eta-path augmented Kn\textbf{O}wledge-grounded \textbf{CHA}tbot, namely \textsc{Mocha}. Given a conversation, our \textsc{Mocha} firstly collects relevant entities as candidates to benefit response generation. Then, \textsc{Mocha} captures the conversation context and generates responses based on an devised encoder-decoder architecture. The encoder compresses the input utterance(s) into a context vector. The decoder is augmented with a pointer gate~\cite{vinyals2015pointer} to decide when to mention an entity and conduct entity reasoning over the pre-collected candidate entities based on meta-path information. Particularly, we define 10 most popular meta-paths observed in our conversation data, and encode them into vectors using metapath2vec approach~\cite{dong2017metapath2vec}. Finally, \textsc{Mocha} is able to generate entities by firstly comparing the context representation with each of the meta-path vectors, and then attending on the candidate entities that follow the most similar meta-path. On the movie corpus we build, our \textsc{Mocha} significantly outperforms the compared models. 

\section{Meta-path Embedding}
Given the entity types from $\mathcal{A}$, a meta-path $\mathcal{P}$ is denoted as $A_1\!\!\to\!\!A_2\!\!\to \!\!\ldots\!\! \to \!\!A_{L+1}$, which defines a semantic relation between types $A_1$ and $A_{L+1}$, where $L$ is the path length.

To augment response generation, we summarize 10 popular meta-paths according to the statistics of the conversation corpus. Each meta-path connects 3 entity types (length of 2), formed by entity mentions in their original order from a same conversation. We depict these 10 meta-paths in the Supplementary Material (Suppl. in short).  
 
After defining these meta-paths, we generate path instances and embed each 2-length instances using a GRU following~\cite{dong2017metapath2vec}. Since a meta-path can produce multiple instances, we aggregate all the instance embeddings into a single vector using a max-pooling operation. As a result, for each of 10 meta-paths, we obtain a vector representation $\mathbf{p}_m$, where $\forall m \in \{1,\ldots,10\}$.

\section{Model: Mocha}
\label{sec:mkb}

Formally, a conversation consists of utterance sequences $\mathbf{x} = \{\mathbf{u}^1, \ldots, \mathbf{u}^{T}\}$, where $T$ is the turn number. Each utterance is a sequence of words. Hence, the input of chatbot is a word sequence $\mathbf{x} = \{x_1,\ldots,x_{N_x}\}$, and the output is a response $\mathbf{y} = \{y_1,\ldots,y_{N_y}\}$, where $N_x$ and $N_y$ are the token numbers. \textsc{Mocha} is a knowledge-grounded chatbot, consisting of three main components, as follows:

\noindent\textbf{\large Entity Collector.} \textsc{Mocha} firstly shortlists a set of conversation-related entities {$E$} from KB, which is detailed described in Suppl. For decoder use, we employ TransE~\cite{transe} to transform the entities into dense embeddings as $\mathbf{E} = \{\mathbf{e}_n$\}, where $\forall n \in \{1,\ldots,N_e\}$. 

\noindent\textbf{\large Context Encoder.} We then embed the input utterances $\mathbf{x}$ using a bi-directional GRU.\footnote{We did not see clear improvements using hierarchical encoder.} The resulting representation at each time step is the concatenation of each direction's state, i.e., $\mathbf{h}_t = [\overleftarrow{\mathbf{h}_t},\overrightarrow{\mathbf{h}_t}]$, which is then fed to the decoder.

\begin{figure}[!t]
\centering 
\includegraphics[height=1.3in]{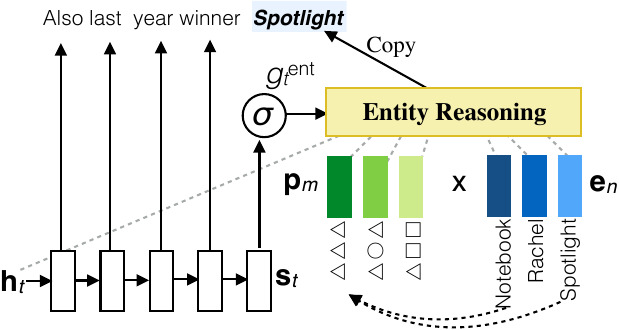}
\caption{Entity-aware Decoder. Best viewed in color.}
\label{fig:decoder}
\end{figure}

\noindent\textbf{\large Entity-aware Decoder.} As shown in Figure~\ref{fig:decoder}, we augment the GRU-based decoder with a gating variable $g_t^{\textnormal{ent}}$~\cite{vinyals2015pointer} that decides whether to generate an entity using $p^{\textnormal{ent}}$ or to omit a general word using $p^{\textnormal{gru}}$. The gate $g_t^{\textnormal{ent}}$ is trained on:
\begin{align*}
g_t^{\textnormal{ent}} &= \sigma(\mathbf{W}_g\mathbf{s}_t)
\end{align*}

When the gate is ``open'', the decoder conducts entity reasoning by taking into account the meta-path information. It firstly approximates how close each meta-path $\mathbf{p}_m$ is to the context $\mathbf{h}_t$, and obtain the attention weights $\boldsymbol{\alpha}_t$ as:
\begin{align*}
\boldsymbol{\alpha}_t \sim \text{exp}(\mathbf{{P}} \mathbf{W}_p \mathbf{h}_t)
\end{align*}
where $\mathbf{P}$ is the matrix consisting of the mata-path vectors $\mathbf{p}_m$. Then, we apply another attention mechanism on $\mathbf{e}_n$ and obtain the corresponding weights $\boldsymbol{\beta}_t$. Intuitively, the generated entity should belong to the ending type ($A_3$) in the attended meta-paths. To do so, we align the entity weights with their corresponding path weights, and multiply the two weights as the output probability. Finally, the decoder generates a candidate entity by: 
\begin{align*}
p^{\textnormal{ent}}(y_t|\mathbf{h}_t, \mathbf{P}, \mathbf{E})=
\begin{cases}
\alpha_{ti}\beta_{tj}, &\text{if } y_t =e_j \text{ and } e_j \mapsto A_i\\
0, &\text{otherwise}
\end{cases}
\end{align*}
When referreing is needed, the decoder directly copies the entity with the highest probability. In this way, the generated response is expected to follow the conversation flow by approximating the context representation $\mathbf{h}_t$ with both meta-path information and candidate entities. 

\section{Experiments}
To validate, we build a movie conversation corpus consisting of roughly 9,000 conversations. We also build a movie KB to link with. The corpus statistics and KB schema are in Suppl.  

\noindent\textbf{\large Compared Models:} (1) \textsc{Attn}: a attention-based Seq2Seq model. (2) \textsc{HRED}~\cite{HRED}: a hierarchical Seq2Seq approach. (3) \textsc{Fact}~\cite{ghazvininejad2017knowledge}: a knowledge-grounded model that consumes textual, unstructured facts. (4) \textsc{GenDS}~\cite{zhu2017gends}: its decoder is also augmented with entity reasoning, but without meta-path taken into account. 

\noindent\textbf{\large Evaluation Metrics:} (1) BLEU-3; (2) Dist-1~\cite{li2016diversity} calculates the ratios of unigrams; (3)(4) Precision and Recall~\cite{zhu2017gends} examines the overlapping on entity mentions in the generated responses.

We constrain the vocabulary to 25,000 words. The embeddings size are 300 and the hidden state vectors are 512. Models are trained using Adam optimizer.

As shown in Table~\ref{exp:compare}, \textsc{Attn} and \textsc{HRED} perform the worst because they are the models that have no access to external KB. We then compare the other three models to find which one(s) utilize(s) knowledge more effectively. Obviously, \textsc{Fact} lags far because it utilizes knowledge described in unstructured text, i.e., \emph{Titanic stars Leonardo as...}. Its disappointing performance suggest that it is more effective to inject structural knowledge into Seq2Seq models.

Remarkably, \textsc{Mocha} and \textsc{GenDS} are the best and second best models. They are the only models with explicit entity reasoning mechanism to generate entities from shortlisted candidates. This proves the necessity of such mechanism for the chatbots. Different from \textsc{GenDS}, \textsc{Mocha} attends on entities based on meta-path information, and prefers those entities that follow the most similar meta-path to the current conversation context. Considering the attention weights on the meta-path vectors, \textsc{Mocha} reduces possibilities of generating incoherent entities when they diverge from the conversation flow.

Drawing on the highest scores achieved by \textsc{Mocha}, it is beneficial to augment response generation using meta-path information. This reveals that the responses generated by \textsc{Mocha} are more informative and satisfactory. For better understanding, we show some generated responses in Suppl. 

\begin{table}[!t]
\begin{tabular}{r|c|c|c|c|c}
\hline
{\textbf{Model}}& \textbf{KB} & \textbf{BLEU-3} & \textbf{Dist-1} & \textbf{Prec.} &  \textbf{Recall} \\\hline
{\textsc{Attn}} & \xmark  & 0.18&0.03&0.14&0.14\\
{\textsc{HRED}} & \xmark & 0.17&0.02&0.13&0.14\\\hline
{\textsc{Fact}} & \cmark & 0.06&0.16&0.13&0.08\\
{\textsc{GenDS}} & \cmark & 0.68&0.16&0.37&0.38\\
{\textsc{Mocha}}& \cmark & \textbf{0.82}&\textbf{0.19}&\textbf{0.48}&\textbf{0.51}\\%
\hline
\end{tabular}
\caption{Experimental Results.}
\label{exp:compare}
\end{table}

\small
\bibliography{aaai2019}
\bibliographystyle{aaai}

\newpage
\section{Supplementary Materials}

In the following, we present the details of our work, including 10 popular meta-paths, entity collector, corpus statistics, KB schema, experimental setup and sampled generated responses.

\section{Popular Meta-paths}
See a conversation on Leonardo DiCaprio in Figure~\ref{fig:example}.

\begin{figure}[!h]
\fbox{\begin{minipage}{0.45\textwidth}
$\mathbf{u}^1$: \emph{Titanic} (film) is really a tragedy.\\
$\mathbf{u}^2$: So is \emph{Romeo and Juliet} (film).\\
$\mathbf{u}^3$: \emph{Total Eclipse} (film) is also a sad story.
\end{minipage}
}
\caption{A conversation example containing a meta-path of F $\to$ F $\to$ F.} 
\label{fig:example}
\end{figure}
By linking the entity mentions to the associate KB, we are able to acquire their corresponding types, i.e., \emph{Titanic} $\mapsto$ type \emph{film}. By connecting the entity types in their original order in the conversation, we obtain the meta-path \emph{film} $\to$ \emph{film} $\to$ \emph{film}.

Based on the conversation data, we extract 10 most popular meta-paths over the entity mentions. Denote that Film (\emph{film}), D (\emph{director}), and A (\emph{actor}). The 10 most popular meta-paths we consider are list in below:
\begin{itemize}
    \item F $\to$ F $\to$ F
    \item F $\to$ A $\to$ A
    \item F $\to$ D $\to$ F
    \item D $\to$ F $\to$ F
    \item D $\to$ F $\to$ D
    \item D $\to$ D $\to$ D
    \item D $\to$ A $\to$ A
    \item A $\to$ A $\to$ A
    \item A $\to$ F $\to$ A
    \item A $\to$ F $\to$ F
\end{itemize}

\section{Entity Collector}

Given a conversation, the first step is to collect the relevant entities from the associate KB to benefit response generation. Typically, a KB organizes knowledge in the form $\{e_h, r, e_t\}$, where $e_h$ and $e_t$ are two entities linked by attribute $r$. While there are numerous facts stored in a KB, only a small fraction of them are related to the given conversation context. 

\begin{figure}[!t]
\centering 
\includegraphics[height=3.3in]{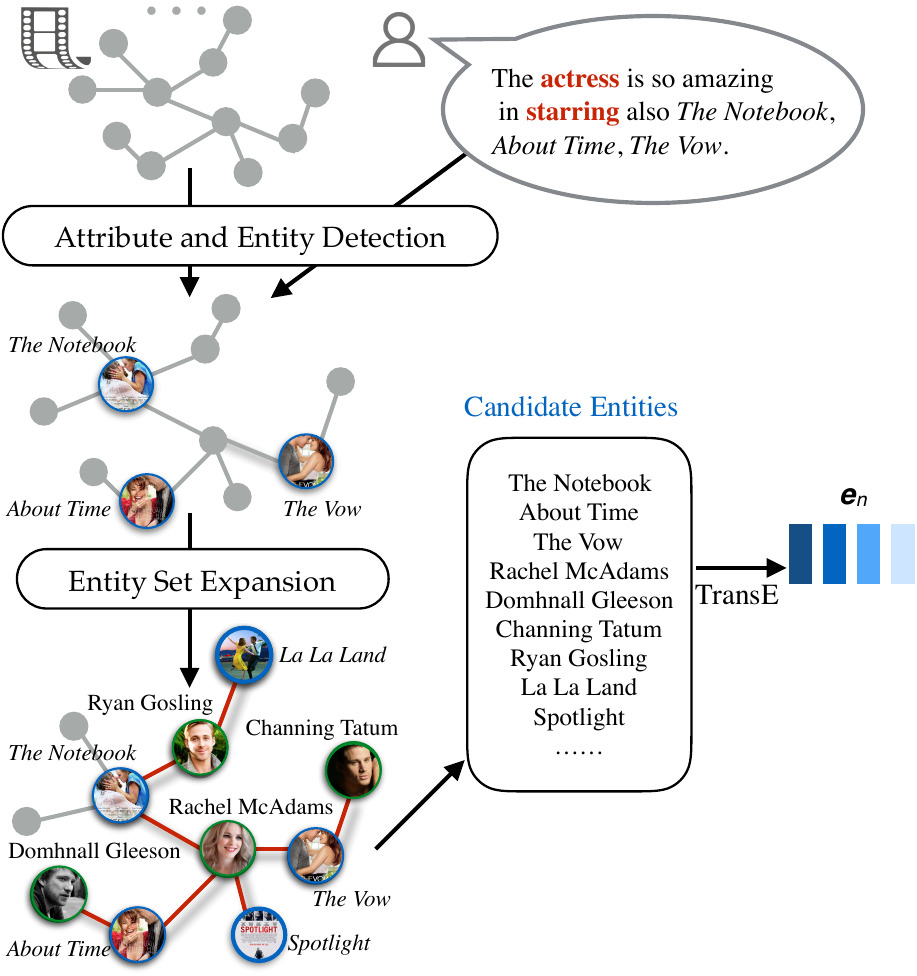}
\caption{Entity Collector.}
\label{fig:knowledge}
\end{figure}

We detect a set of entities mentioned in a conversation using entity linking techniques~\cite{shen2015entity}. To encourage more diverse and richer conversation content, we further expand the entity set as follows. We treat the detected entities as seeds (three circled clear pictures in the middle of Figure~\ref{fig:knowledge}) and add their neighboring entities within 2-hops. All detected and added entities serve as the entity candidates to be selected in response generation.

As a result, we shortlist a set of conversation-related entity candidates {$E$}. For generalization, we employ the knowledge graph embedding model TransE~\cite{transe} to transform them into corresponding dense embeddings, denoted as $\mathbf{e}_n$, where $\forall n \in \{1,\cdots,N_e\}$. These entity embeddings are then fed to the decoder to serve as candidates in entity-aware response generation when referring is needed.

\section{Corpus Statistics}

To validate the proposed model, we build a new multi-round conversation corpus, \textsc{Bili-film}, which is collected from \textsc{BiliBili}, one of the largest Chinese video sharing and discussion platform.\footnote{\url{https://www.bilibili.com/v/cinephile/}} 
On \textsc{bilibili}, users publish movie-related videos, and discuss the videos by leaving new comments or responding to existing comments. We define a seed set of 20 active publishers to crawl under their videos the conversations between two users. 
We filter out the conversations that are longer than four speaker turns. The statistics of the corpus is presented in Table~\ref{table:bili-film}. {The corpus will be released to the public.}

\section{Knowledge Base Schema}
We build our KB based on \textsc{zhishi.me}~\cite{zhishi}, the largest Chinese knowledge base comprising comprehensive knowledge from three Chinese encyclopedias: Wikipedia Chinese version, Baidu Baike, and Hudong Baike.\footnote{\url{https://baike.baidu.com}, \url{https://www.hudong.com}} Despite being a general KB, \textsc{zhishi.me} has the largest coverage in movie domain compared to others.

\begin{figure}[!t]
\centering 
\includegraphics[height=1.2in]{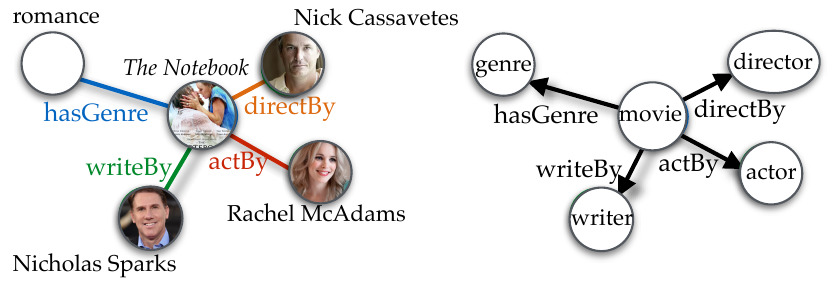}
\caption{Left: A Snapchat Instance of Our KB. Right: The Schema of Our KB.}
\label{fig:schema}
\end{figure}

We demonstrate the schema of our KB in Figure~\ref{fig:schema}. To acquire it, we extract from \textsc{zhishi.me} the triples containing either the attribute \emph{actBy} or \emph{directBy}. This assumes to acquire all the films in it. As common practice, we add inverse attributes (i.e., \emph{actBy}$^{-1}$), and re-collect triples about these films according to five attributes: \emph{hasAlias}, \emph{directBy}, \emph{actBy}, \emph{writeBy},\footnote{Some films are adapted from books, for example, the series of Harry Potter. In this case, we consider the writer of the book.} and \emph{hasGenre}. Correspondingly, there are five types of entities in the KB, i.e., film, director, actor (actress), writer, and genre.

\begin{table}[!t]
\centering
\small
\begin{tabular}{|r|r|}
\hline
{Number of Total Conversations} & {8,768}\\
{Average Conversation Turns} & {3.6} \\
{Average Tokens Per Turn} & {27.8}\\\hline
{Number of Covered Films} & {162} \\
{Number of Covered Film Stars} & {239} \\
{Average Entities Per Turn} & {2.4} \\
{Unique Entities Per Conversation} & {3.1} \\\hline
\end{tabular}
\caption{Statistics of Corpus \textsc{Bili-film}}\label{table:bili-film}
\end{table}

\section{Experimental Setup}
All the compared models are implemented by TensorFlow~\cite{tensorflow}. To setup the experiments, the corpus is tokenized by Jieba Chinese word segmenter.\footnote{\url{https://github.com/fxsjy/jieba}} For TransE we utilize to encode our MKB, we adopt the implementation from KB2E~\cite{kb2e}.\footnote{\url{https://github.com/thunlp/KB2E}} We constrain the vocabulary to 25,000 words. The word embeddings are initialized with FastText vectors~\cite{fasttext}, and fine-tuned during training. The embeddings size are 300 and the hidden state vectors are 512. We set the mini-batch size as 32, and the learning rate to be 0.001 initially, which is decayed exponentially during training. We also clip gradients with norms larger than 0.5. Models are trained using Adam optimizer~\cite{adam}.

\section{Generated Responses}
The experiments presented in the submission are conducted using automatic evaluations. For better understanding, we also sample some generated responses to manually judge the model performances, which are presented in Table~\ref{table:case} in the next page.

\begin{table*}[!t]
\centering
\begin{adjustbox}{max width=\textwidth} 
\begin{tabular}{|l|l|l|}
\hline
{\cellcolor{gray!25}{Film} }& {\cellcolor{gray!25}{\begin{CJK*}{UTF8}{gkai}恋恋笔记本\end{CJK*}} }& {\cellcolor{gray!25}{The Notebook} }\\\hline
\textbf{Input} & $\mathbf{u}^1$: \begin{CJK*}{UTF8}{gkai}女主演过\underline{贱女孩}, 相当好看\end{CJK*} & $\mathbf{u}^1$: The actress also stars in \underline{\emph{Mean Girls}}, amazing!  \\
& $\mathbf{u}^2$: \begin{CJK*}{UTF8}{gkai}推荐你看\underline{时空恋旅人}，很感动\end{CJK*} & $\mathbf{u}^2$: I recommend \underline{\emph{About Time}}, really touching!  \\\hline
\textbf{\textsc{Attn}}& \begin{CJK*}{UTF8}{gkai}你没看懂\end{CJK*}  &  You didn't understand.\\
\textbf{\textsc{HRED}}& \begin{CJK*}{UTF8}{gkai}好看好看好看\end{CJK*}  & It is good good good.\\
\textbf{\textsc{Fact}}& \begin{CJK*}{UTF8}{gkai}我很喜欢\end{CJK*} & I like it.\\
\textbf{\textsc{GenDS}}& \begin{CJK*}{UTF8}{gkai}是\underline{瑞秋}\end{CJK*} & It's \underline{Rachel}.\\
\textbf{\textsc{Mocha}}&
\begin{CJK*}{UTF8}{gkai}女主演了很多，还有\underline{爱乐之城}\end{CJK*} & The actress stars in a lot of films, such as \emph{\underline{La La Land}}.\\
\hline
{\cellcolor{gray!25}{Film} }& {\cellcolor{gray!25}{\begin{CJK*}{UTF8}{gkai}美人鱼\end{CJK*}} }& {\cellcolor{gray!25}{Mermaid} }\\\hline
\textbf{Input} & $\mathbf{u}^1$: \begin{CJK*}{UTF8}{gkai}女主的演技太让人受不了了\end{CJK*} & $\mathbf{u}^1$: The actress's acting is intolerable.  \\
& $\mathbf{u}^2$: \begin{CJK*}{UTF8}{gkai}只有\underline{罗志祥}的演技能看\end{CJK*} & $\mathbf{u}^2$: Only \underline{Zhixiang Luo}'s acting skill is above average. \\\hline
\textbf{\textsc{Attn}}& \begin{CJK*}{UTF8}{gkai}确实很好\end{CJK*}  &  Really good.\\
\textbf{\textsc{HRED}}& \begin{CJK*}{UTF8}{gkai}女主不好看\end{CJK*}  & The actress is not beautiful.\\
\textbf{\textsc{Fact}}& \begin{CJK*}{UTF8}{gkai}电影好\end{CJK*} & The film is good.\\
\textbf{\textsc{GenDS}}& 
\begin{CJK*}{UTF8}{gkai}\underline{周星驰}的\underline{功夫}好\end{CJK*} &\underline{Stephen Chow}'s \emph{\underline{Kung Fu}} is good.\\
\textbf{\textsc{Mocha}}& \begin{CJK*}{UTF8}{gkai}\underline{美人鱼}中\underline{罗志祥}不错\end{CJK*} & In \emph{\underline{Mermaid}}, {\underline{Zhixiang Luo}} is really good. \\
\hline

{\cellcolor{gray!25}{Film} }& {\cellcolor{gray!25}{\begin{CJK*}{UTF8}{gkai}敦刻尔克\end{CJK*}} }& {\cellcolor{gray!25}{Dunkirk} }\\\hline
\textbf{Input} & $\mathbf{u}^1$: \begin{CJK*}{UTF8}{gkai}\underline{敦刻尔克}是\underline{诺兰}最好的电影\end{CJK*} & $\mathbf{u}^1$: \emph{\underline{Dunkirk}} is the best film of \underline{Nolan}.  \\
& $\mathbf{u}^2$: \begin{CJK*}{UTF8}{gkai}有点夸张，只是\underline{诺兰}的一次尝试\end{CJK*} & $\mathbf{u}^2$: It's overstated. I think this is only a \underline{Nolan}'s experiment. \\\hline
\textbf{\textsc{Attn}}& \begin{CJK*}{UTF8}{gkai}我觉得不好看\end{CJK*}  &  I don't like it.\\
\textbf{\textsc{HRED}}& \begin{CJK*}{UTF8}{gkai}不好\end{CJK*}  & Bad.\\
\textbf{\textsc{Fact}}& \begin{CJK*}{UTF8}{gkai}漫威电影\end{CJK*} & Marvel's film.\\
\textbf{\textsc{GenDS}}& 
\begin{CJK*}{UTF8}{gkai}\underline{诺兰}黑\end{CJK*} &\underline{Nolan}'s films are dark. \\
\textbf{\textsc{Mocha}}& \begin{CJK*}{UTF8}{gkai}\underline{敦刻尔克}很不错，\underline{诺兰}都喜欢\end{CJK*} & \emph{\underline{Dunkirk}} is superb, and I like all {\underline{Nolan}}'s films.\\
\hline
\end{tabular}
\end{adjustbox}
\caption{Sampled Generated Responses. Underlined words are entities. Input utterances are shorten due to limit space. We translate the original conversation into English for better understanding.}\label{table:case}
\end{table*}

\end{document}